\documentclass[sigconf]{acmart}

\AtBeginDocument{%
  }

\copyrightyear{2025}
\acmYear{2025}
\setcopyright{cc}
\setcctype{by}
\acmConference[GECCO '25]{Genetic and Evolutionary Computation Conference}{July 14--18, 2025}{Malaga, Spain}
\acmBooktitle{Genetic and Evolutionary Computation Conference (GECCO '25), July 14--18, 2025, Malaga, Spain}\acmDOI{10.1145/3712256.3726375}
\acmISBN{979-8-4007-1465-8/2025/07}

\usepackage{enumitem}
\usepackage{subcaption}
\usepackage{bbm}
\usepackage{multirow}
\usepackage[ruled,vlined]{algorithm2e}
\SetAlgorithmName{Algorithm}{Algorithm}{Algorithm}
\SetKwProg{Proc}{Procedure}{:}{}
\SetAlgoSkip{} 
\usepackage{soul}

\begin{document}

\title{Analysis of Memory-Runtime Trade-offs in Caching Strategies for Genetic Programming Symbolic Regression}

\author{Jiaming Shi}
\email{shij@u.nus.edu}
\orcid{0009-0005-2776-8938}
\affiliation{%
 \institution{National University of Singapore}
 \department{Department of Electrical and Computer Engineering}
 \city{Singapore}
 \country{Singapore}
}

\author{Kei Sen Fong}
\email{fongkeisen@u.nus.edu}
\orcid{0009-0000-4135-4858}
\affiliation{%
 \institution{National University of Singapore}
 \department{Department of Electrical and Computer Engineering}
 \city{Singapore}
 \country{Singapore}
}

\author{Mehul Motani}
\email{motani@nus.edu.sg}
\orcid{0000-0003-3262-0207}
\affiliation{%
 \institution{National University of Singapore}
 \department{Department of Electrical and Computer Engineering, Institute of Data Science, N.1 Institute for Health, Institute for Digital Medicine}
 \city{Singapore}
 \country{Singapore}
}


\begin{abstract}
Genetic Programming Symbolic Regression (GPSR) generates mathematical expressions to model input-output relationships using an evolutionary process. A significant challenge in GPSR lies in the repeated evaluation of entire expressions or their sub-expression, which inflates computational runtime. To address this inefficiency, caching mechanisms have been employed to reduce redundant computations. However, prior studies predominantly employ a single caching strategy, offering limited insights into their comparative performance or memory-runtime trade-offs. In this paper, we present a comprehensive analysis of caching mechanisms for GPSR on synthetic and real-world datasets. We also include an empirical study of key-value usage frequencies under an infinitely large cache, offering insights into optimal cache sizing. Furthermore, we provide actionable guidelines for configuring caching strategies based on computational and memory constraints. Our findings indicate that complex caching mechanisms necessitate a minimum cache size to achieve computational time reductions. Conversely, lightweight caching strategies, such as Least Recently Used (LRU) and, notably, First-In-First-Out (FIFO), can significantly decrease computation time for fitness evaluations, which are a substantial component of the overall runtime.

\end{abstract}




\maketitle

\section{INTRODUCTION}
Symbolic regression (SR) is a robust machine learning technique designed to model input-output data by autonomously identifying both the structural form and parameters of mathematical expressions that accurately capture the underlying data relationships \cite{banzhaf2024nature, fong2024multi, bakurov2022genetic, geiger2023down, de2023reducing}. SR is particularly advantageous when the objective is to represent a process through concise mathematical formulations. This methodology has demonstrated significant utility across diverse disciplines, including physics, finance, biology, material, and various engineering sectors \cite{jameer2024stability, ahmadi2024ai, chen2021empirical,fong2024symbolic}. 

\begin{figure}[!t]
  \centering
  \includegraphics[width=\linewidth]{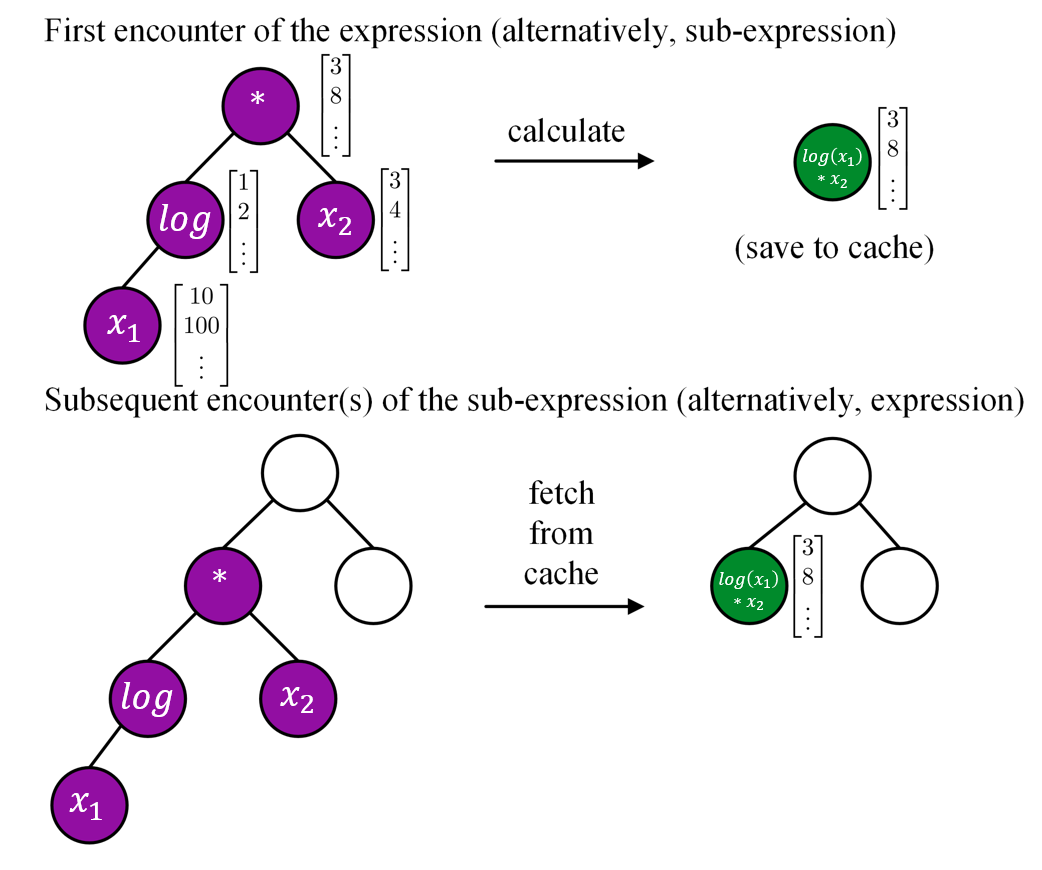}
  \caption{Illustration of cache replacement in GPSR. Here, after the first encounter of $log(x_1)*x_2$ (either as a full expression or sub-expression), future encounters will no longer require calculation and would instead be fetched from cache.}
  \label{cachepic}
  \vspace{-4mm}
\end{figure}

Numerous GPSR tools have been developed to support these applications. 
In this work, we selected gplearn \cite{stephens2016genetic}, a widely used Python-based SR library that closely adheres to Koza’s original genetic programming (GP) framework \cite{koza1992genetic}. This choice ensures that our proposed mechanisms are evaluated in a setting that reflects fundamental GP principles, making our findings more broadly applicable. In contrast, many modern SR tools incorporate additional heuristics, which could introduce confounding factors and limit the generality of our analysis.

During the training process of SR, we often observe that the majority of GPSR's runtime is not spent on generating suitable candidate expressions, but rather on calculating the fitness of these candidate expressions for the dataset. Furthermore, on larger datasets, this computation time often accounts for half or even more of the total runtime, making it one of the primary bottlenecks in GPSR's performance. Since GPSR relies on evolutionary techniques as a mechanism to search for the optimal expression for the dataset, it evaluates and ranks the candidate expressions based on their fitness at the end of each training generation. Then a subset of these expressions is retained for the next generation of training. Many of the new expressions in the next generation are typically modifications of the high-fitness expressions from the previous generation. This results in significant computational redundancy \cite{imai2024inexact}, as many calculations performed in one generation are repeated in subsequent generations.

To address this issue, several studies have proposed incorporating caching mechanisms or employing similar concepts to store intermediate computation results in SR algorithms \cite{wong2008removing}. For instance, \cite{350024} suggests storing all equations in one large graph. Similarly, \cite{5949751} applies caching in linear genetic programming, which, although not identical to genetic programming, provides valuable insights for this field. \cite{4631158} proposes SCHEME (Subtree Caching using a Hashing for Equivalence MEthod), a method that uses hashing to estimate algebraic equivalence between subtrees and thus reduces node evaluations. \cite{keijzer2004alternatives, wong2008removing} mention several caching algorithms, such as the use of LRU. However, these do not provide a detailed comparison of different caching strategies and the memory-runtime trade-offs.

As shown in Figure \ref{cachepic}, if the same expression (or sub-expression) appear again, the cached results can be reused instead of recalculating them again. Our experiment extends the capabilities of the GPSR by incorporating caching strategies, providing a comprehensive comparison of their performance impacts from multiple perspectives, such as different caching strategies, varying cache sizes, and their effects on runtime, memory usage, and other factors.

\noindent The main \textbf{contributions} of this paper are as follows: 
\begin{enumerate}[nosep, wide]
\item We enhanced GPSR by incorporating a caching mechanism, which significantly reduces the fitness computation runtime.
\item We conducted a comprehensive comparison of various caching strategies and cache sizes, analyzing the impact on training runtime, memory consumption and their trade-off.
\item We investigated whether periodically proactive cache clearance can accelerate training, providing detailed comparisons of how different clearance intervals affect the runtime.
\item We conducted access count tests under infinite cache conditions and proposed to use the concept of RAM hour to further quantify the memory-runtime trade-offs, and provided reference guidelines for practitioners to determine appropriate cache sizes and configuration strategies.
\end{enumerate}

\section{RELATED WORK}
\subsection{Symbolic Regression}
SR is a machine learning and data modeling approach designed to identify mathematical expressions that describe the relationships between variables based on data. Unlike traditional regression methods (e.g., linear regression or polynomial regression), SR does not assume a predefined model structure. Instead, it employs algorithms, such as GP, to automatically search for the optimal mathematical expression. The output of SR, being explicit mathematical formulas, is inherently interpretable and easy for humans to understand.

There are numerous advancements in SR, with some notable branches. PySR, for example, has significantly enhanced performance by using a multi-population evolutionary approach \cite{cranmer2023interpretable}. MetaSR focuses on optimizing fitness evaluation by utilizing SR itself to discover new fitness measures, which may be complex combinations of existing base measures \cite{fong2024metasr}. TaylorGP enhances SR by leveraging Taylor polynomials to approximate symbolic equations and extract key features, such as low-order polynomial discrimination, variable separability, boundary behavior, monotonicity, and parity \cite{he2022taylor}. These techniques improve the performance of SR. In this work, we chose gplearn implementation of SR \cite{stephens2016genetic}, which adheres closely to Koza’s original genetic programming (GP) framework \cite{koza1992genetic}. This choice ensures that our proposed mechanisms are evaluated in a setting that reflects fundamental GP principles instead of incorporating the additional heuristics of the advanced methods mentioned above, making our findings more broadly applicable.

\subsection{Cache Strategies}
When using caching, the average required time can be expressed as a contribution of a few components \cite{smith1987design}: $T = m \times T_m + T_h + E$, where \( m\): cache miss rate (i.e., \(1 - \text{hit rate} \)), \( T_m \): time required when a cache miss occurs, \( T_h \): time required to access the cache, \( E \): time required for other secondary effects.

In general, the performance of a cache primarily depends on two key metrics: latency and hit rate. Additionally, some secondary factors can also influence cache performance. When the caching mechanism (e.g., storing cached results in a dictionary) is fixed, latency is often difficult to significantly change under the same cache size and other conditions. However, we can optimize the overall cache performance by improving the hit rate through appropriate cache policies.

\subsubsection{Common Practical Cache Policies} Below are four commonly used cache policies which are used in real-world applications:
\begin{enumerate}[nosep, wide, leftmargin=16pt, itemindent=-10pt]
\item Least Recently Used (LRU): Evicts the least recently used pages based on past access patterns as an approximate prediction of future access patterns.
\item Least Frequently Used (LFU): Evicts the least frequently accessed pages, suitable for specific workloads.
\item First In, First Out (FIFO): Simply evicts the pages that entered the cache the earliest.
\item Random Replacement (RR): Randomly selects pages to evict, commonly used in certain hardware systems.
\end{enumerate}

\subsubsection{Other Cache Policies}
Bélády's algorithm \cite{belady1966study} is another well-known strategy that evicts the data element that will not be used for the longest time in the future. Also known as the Optimal Page Replacement Algorithm, it is theoretically the most optimal caching strategy. However, as a foresight-based strategy, it cannot be practically implemented because it requires complete knowledge of future access sequences, which is not practical in GPSR since the evolutionary process is innately sequential.

\section{METHODOLOGY}

\begin{algorithm}[!t]
\caption{Integration of Cache in GP’s \textit{execute} Function}
\label{cache_in_execute}
\SetAlgoLined
\DontPrintSemicolon
\SetKwBlock{DoParallel}{do in parallel}{end}
\KwIn{$X$, $self.program$}
\KwOut{$y\_pred$}
$stack = [ ]$\;
$index \gets 0$\;

\While{$index < len(self.program)$}{
    $node \gets self.program[index]$\;
    \If{$node \in cache$}{
        $intermediate\_result \gets cache[node\_expression]$\;
        \tcc{Get cache record.}
        $stack[-1].append(intermediate\_result)$\;        
        $index \gets index + len(node) - 1$\;
    }
    \Else{
        $stack[-1].append(node)$\;
    }
    \tcc{If the cache has already recorded the computation result of the node, directly use the cache record. Otherwise, traverse step by step from the root node to the leaf node.}
    \While{$stack[-1] \text{is a node with a depth of 1}$}{
    $intermediate\_result \gets compute\_node(stack[-1])$\;
    $cache[node\_expression] \gets intermediate\_result$\;
    \tcc{Add cache record. The depth of node will become smaller and smaller as it is calculated layer by layer from leaf node to root node, but node\_expression records the complete sub-expression}
    \If{$len(stack) != 1$}{
        $stack.pop()$\;
        $stack[-1].append(intermediate\_result)$\;
    }
    \Else{
        $y\_pred \gets intermediate\_result$\;
        \Return{$y\_pred$}
    }
    }
    $index \gets index + 1$\;
}
\end{algorithm}

Our integration of cache into GPSR is shown in Algorithm \ref{cache_in_execute}. In the implementation of GPSR by gplearn, raw fitness is calculated based on the candidate expressions. The \textit{execute} function traverses from the root node gradually to the leaf nodes, recording the structure in a list. After reaching the leaf nodes, numerical computations proceed from the leaf nodes step by step upwards to the root node to obtain the final result.

The cache mechanism, on the other hand, during the traversal from the root node to the leaf nodes, checks whether the computation result of a sub-expression is already cached. If it is, the cached result is directly used to replace the entire sub-expression; otherwise, the traversal continues downward. Once the numerical computation progresses from the bottom up to that node, the computation result of this sub-expression is added to the cache. The steps in Algorithm \ref{cache_in_execute} illustrates this process in pseudo code.

The modification has a minimal impact on the overall gplearn implementation, with the main changes confined to the \textit{execute} function. This makes it easy to integrate with other future modifications and mechanisms to gplearn without causing conflicts.
For different caching algorithms, we only need to implement a cache class or use a caching library. The main parts of gplearn's code remain generic, which also facilitates detailed and fair comparisons of various types of caches.

In this work, we primarily utilized the cachetools library \cite{cachetools} to implement different cache strategies, enabling a fair comparison of the acceleration effects of LRU, FIFO, LFU and RR caching algorithms on GPSR.

\section{RESULTS AND DISCUSSION}

\subsection{Real-World Dataset Experiments}

\begin{table}[t]
\centering
\caption{Tuned Parameters for gplearn from SRBench \cite{srbench}}
{%
\begin{tabular}{|l|l|}
\hline
\textbf{Parameter} & \textbf{Value} \\ \hline
Generations & 500 \\ \hline
Initial Method & half and half \\ \hline
Low Memory & False \\ \hline
Max Samples & 1.0 \\ \hline
Function Set & add, sub, mul, div \\ \hline
Metric & mean absolute error \\ \hline
Jobs & 1 \\ \hline
Crossover Probability & 0.9 \\ \hline
Hoist Mutation Probability & 0.01 \\ \hline
Point Mutation Probability & 0.01 \\ \hline
Point Replace Probability & 0.05 \\ \hline
Subtree Mutation Probability & 0.01 \\ \hline
Parsimony Coefficient & 0.001 \\ \hline
Population Size & 1000 \\ \hline
Stopping Criteria & 0.0 \\ \hline
Tournament Size & 20 \\ \hline
Verbose & 0 \\ \hline
Warm Start & False \\ \hline
\end{tabular}%
}
\label{Parameter Settings for GP}
\end{table}

\begin{table*}[!t]
\fontsize{9pt}{9.5pt}\selectfont
\centering
\caption{Results for Original Implementation of GPSR by gplearn}
\begin{tabular}{l|llll}
\bottomrule
Dataset & Overall Runtime ($s$) & \textit{execute} Runtime ($s$) & Average RAM (MB) & Maximum RAM (MB)\\ \toprule \bottomrule
344\_mv & 657.3 ± 24.8 & 330.2 ± 20.6 & 248.5 ± 31.1 & 403.7 ± 45.3 \\
215\_2dplanes & 918.4 ± 230.8 & 591.7 ± 223.3 & 207.7 ± 5.0 & 267.9 ± 16.8 \\
1203\_BNG\_pwLinear & 2094.2 ± 9.3 & 1445.1 ± 9.7 & 261.5 ± 0.6 & 344.9 ± 0.8 \\ \toprule \bottomrule
\end{tabular}
\label{Results for original gplearn}
\end{table*}

\begin{table}[!b]
\fontsize{9pt}{9.5pt}\selectfont
\centering
\caption{Test Metrics: $R^2$ and MAE for Datasets}
\begin{tabular}{l|ll}
\bottomrule
Dataset & Test $R^2$ & Test MAE \\ \toprule \bottomrule
344\_mv & 0.965 & 0.658 \\
215\_2dplanes & 0.894 & 1.102 \\
1203\_BNG\_pwLinear & 0.568 & 2.109 \\ \toprule \bottomrule
\end{tabular}
\label{Test R^2 & MAE}
\end{table}

We study three datasets from Penn Machine Learning Benchmarks (PMLB) \cite{romano2021pmlb}: 344\_mv, 215\_2dplanes, and 1203\_BNG\_pwLinear, with dimensions (40768, 10), (40768, 10) and (177147, 10), respectively. For each dataset, the random state was set from 0 to 7 to create training and test sets in the ratio of 75:25. The stopping criteria is set to 0 to ensure that the number of training epochs is exactly 500. When using the same seed, this also guarantees consistent computational load, making it easier to compare the time savings achieved by different strategies. The runtimes are measured on a clean Google Cloud Platform c3-highcpu-22 instance and aggregated over the random states for each dataset.

The mean and standard deviation for each dataset were calculated using values between the 75th and 25th percentiles to remove anomalous readings in a fair manner. All other hyperparameters were set in accordance with the default configurations in SRBench \cite{srbench}, with the specific parameter settings as shown in Table \ref{Parameter Settings for GP}.

We use the original gplearn as the control group. Table \ref{Results for original gplearn} presents the overall (i.e., total) runtime, \textit{execute} runtime and RAM usage required by the original gplearn, serving as a baseline for comparison with gplearn employing different caching strategies.

We also recorded the performance on the test set under these conditions as a reference, which are similar to the results obtained from SRBench \cite{srbench} since we used their tuned hyperparameters and caching does not affect the prediction performance of the algorithm. These are recorded in Table \ref{Test R^2 & MAE} which presents the average $R^2$ and MAE values across different random seeds for the datasets.

\subsubsection{Runtime Analysis}

For GPSR (i.e., gplearn), the computation time for raw fitness during each generation's training accounts for a significant proportion, and the \textit{execute} function is responsible for calculating raw fitness. We provide a detailed comparison of the overall training runtime and the cumulative runtime of the \textit{execute} function under different cache size and cache strategies (see Tables \ref{Runtime for dataset 344_mv}, \ref{Runtime for dataset 215_2dplanes}, \ref{Runtime for dataset 1203_BNG_pwLinear}).
We use the cProfile library to record the runtime in detail, and all results are measured in seconds.

For the LRU and FIFO caching strategies, as the cache size increases, both the overall runtime and the cumulative runtime of the \textit{execute} function decrease.  However, this reduction in time exhibits diminishing returns.  When the cache size becomes sufficiently large, further increases in cache size bring diminishing improvements to the overall runtime, eventually becoming negligible. Moreover, in most cases, LRU generally provides slightly better runtime improvements than FIFO, except when the cache size is excessively large, where FIFO performs slightly better than LRU. For the original gplearn, the cumulative runtime of the \textit{execute} function accounts for nearly 50\% or even more of the overall runtime, making it the primary bottleneck in gplearn's performance. However, when the cache size is sufficiently large, the runtime of the \textit{execute} function accounts for only about 10\% of the overall runtime. In some cases, such as the 344\_mv dataset, with a cache size of 10k and the LRU caching strategy, this proportion drops even below 5\%. This indicates that with caching acceleration, the computation of fitness is no longer the primary bottleneck in gplearn's performance. 

On the other hand, for the LFU caching strategy, although certain cache sizes do reduce runtime compared to the original gplearn, they consistently underperform compared to the LRU and FIFO caching strategies under the same conditions. Furthermore, when the cache size becomes excessively large, runtime increases significantly, and in some cases, the LFU caching strategy even results in longer runtimes than the original gplearn. This suggests that the LFU caching strategy may not be suitable for accelerating GPSR. RR also demonstrated relatively weak performance, which is intuitively aligned with expectations. Surprisingly, when the cache size is small (e.g. 100 or 500), RR performs almost as well as LRU.

\begin{table*}[!t]
\fontsize{9pt}{9.5pt}\selectfont
\centering
\caption{Runtime (Measured in $s$) for Dataset 344\_mv}
{
\resizebox{\textwidth}{!}{
\begin{tabular}{l|ll|ll|ll|ll}
\bottomrule
~ & \multicolumn{2}{c|}{LRU} & \multicolumn{2}{c|}{FIFO} & \multicolumn{2}{c|}{LFU} & \multicolumn{2}{c}{RR} \\
Cache size & Overall & \textit{execute} & Overall & \textit{execute} & Overall & \textit{execute} & Overall & \textit{execute} \\ \toprule \bottomrule
100 & 280.6 ± 10.2 & 59.4 ± 12.1  & 298.9 ± 3.1 & 75.5 ± 6.0  & 535.8 ± 48.5 & 303.7 ± 49.6  & 286.2 ± 2.9 & 59.7 ± 7.5  \\ 
500 & 242.4 ± 4.5 & 24.4 ± 6.0  & 248.1 ± 5.7 & 30.1 ± 7.6  & 357.1 ± 7.9 & 133.3 ± 4.2  & 247.2 ± 2.7 & 25.1 ± 5.2  \\ 
1k & 231.6 ± 2.1 & 17.7 ± 3.1  & 235.5 ± 2.6 & 21.6 ± 4.9  & 272.4 ± 11.2 & 54.3 ± 7.3  & 239.2 ± 1.7 & 19.1 ± 3.2  \\ 
5k & 230.9 ± 1.5 & 11.7 ± 0.5  & 229.9 ± 1.7 & 11.9 ± 1.0  & 242.0 ± 0.8 & 24.4 ± 1.8  & 233.9 ± 3.2 & 15.1 ± 1.2  \\ 
10k & 231.1 ± 1.1 & 11.1 ± 0.2  & 228.1 ± 2.4 & 10.0 ± 0.5  & 251.7 ± 0.6 & 31.5 ± 2.7  & 236.5 ± 3.8 & 16.8 ± 1.0  \\ 
50k & 232.9 ± 1.4 & 12.0 ± 0.1  & 228.9 ± 3.2 & 10.6 ± 0.1  & 298.1 ± 5.6 & 18.0 ± 6.8  & 249.9 ± 4.4 & 12.3 ± 1.7  \\ 
100k & 230.2 ± 1.2 & 10.4 ± 0.1  & 226.2 ± 3.2 & 8.9 ± 0.1  & 297.1 ± 7.8 & 9.9 ± 0.1  & 246.9 ± 4.7 & 8.6 ± 0.1  \\  \toprule \bottomrule
\end{tabular}
}
\label{Runtime for dataset 344_mv}}
\end{table*}

\begin{table*}[!t]
\fontsize{9pt}{9.5pt}\selectfont
\centering
\caption{Runtime (Measured in $s$) for Dataset 215\_2dplanes}
\resizebox{\textwidth}{!}{
\begin{tabular}{l|ll|ll|ll|ll}
\bottomrule
~ & \multicolumn{2}{c|}{LRU} & \multicolumn{2}{c|}{FIFO} & \multicolumn{2}{c|}{LFU} & \multicolumn{2}{c}{RR} \\
Cache size & Overall & \textit{execute} & Overall & \textit{execute} & Overall & \textit{execute} & Overall & \textit{execute} \\ \toprule \bottomrule
100 & 489.9 ± 91.0 & 249.1 ± 83.8  & 514.9 ± 104.7 & 273.3 ± 98.0  & 1058.9 ± 320.8 & 819.0 ± 314.3  & 516.2 ± 100.4 & 268.3 ± 92.6  \\ 
500 & 468.8 ± 90.3 & 217.2 ± 81.3  & 479.8 ± 93.3 & 226.0 ± 84.1  & 748.2 ± 302.9 & 512.7 ± 295.7  & 475.8 ± 102.7 & 218.5 ± 92.2  \\ 
1k & 424.0 ± 99.9 & 178.9 ± 88.5  & 441.1 ± 94.0 & 192.2 ± 84.7  & 703.0 ± 241.6 & 468.5 ± 234.4  & 441.2 ± 110.7 & 189.2 ± 97.0  \\ 
5k & 343.0 ± 89.7 & 109.6 ± 76.9  & 356.7 ± 91.6 & 118.5 ± 79.2  & 772.8 ± 279.9 & 541.6 ± 272.0  & 394.1 ± 115.2 & 153.5 ± 101.2  \\ 
10k & 317.1 ± 69.3 & 85.9 ± 60.2  & 327.0 ± 77.4 & 93.6 ± 65.9  & 990.9 ± 538.1 & 758.1 ± 528.0  & 412.2 ± 131.4 & 173.5 ± 118.2  \\ 
50k & 299.6 ± 52.6 & 64.5 ± 43.4  & 296.7 ± 52.1 & 63.2 ± 43.6  & 1524.7 ± 970.1 & 1227.7 ± 963.2  & 748.3 ± 384.3 & 477.2 ± 366.6  \\ 
100k & 297.7 ± 51.6 & 61.3 ± 41.6  & 293.3 ± 48.5 & 59.8 ± 40.9  & 2368.2 ± 1631.7 & 1870.0 ± 1608.0  & 1154.4 ± 727.7 & 794.7 ± 696.6  \\  \toprule \bottomrule
\end{tabular}
}
\label{Runtime for dataset 215_2dplanes}
\end{table*}

\begin{table*}[!t]
\fontsize{9pt}{9.5pt}\selectfont
\centering
\caption{Runtime (Measured in $s$) for Dataset 1203\_BNG\_pwLinear}
{
\resizebox{\textwidth}{!}{
\begin{tabular}{l|ll|ll|ll|ll}
\bottomrule
~ & \multicolumn{2}{c|}{LRU} & \multicolumn{2}{c|}{FIFO} & \multicolumn{2}{c|}{LFU} & \multicolumn{2}{c}{RR} \\
Cache size & Overall & \textit{execute} & Overall & \textit{execute} & Overall & \textit{execute} & Overall & \textit{execute} \\ \toprule \bottomrule
100 & 849.9 ± 3.6 & 362.7 ± 2.3  & 887.1 ± 5.4 & 398.5 ± 2.8  & 1635.2 ± 43.7 & 1211.3 ± 45.7  & 851.1 ± 3.2 & 368.0 ± 3.0  \\ 
500 & 588.9 ± 2.8 & 143.8 ± 1.7  & 667.2 ± 3.9 & 208.1 ± 1.6  & 954.5 ± 3.4 & 536.8 ± 4.4  & 616.5 ± 2.8 & 160.8 ± 2.2  \\ 
1k & 503.1 ± 3.3 & 73.9 ± 1.2  & 563.7 ± 3.3 & 122.6 ± 1.2  & 803.7 ± 3.8 & 381.3 ± 4.3  & 535.0 ± 4.0 & 94.3 ± 1.9  \\ 
5k & 462.0 ± 2.4 & 39.0 ± 0.8  & 468.9 ± 2.4 & 43.5 ± 0.7  & 605.0 ± 8.1 & 182.1 ± 8.5  & 482.2 ± 3.5 & 52.3 ± 1.4  \\ 
10k & 458.1 ± 2.6 & 35.2 ± 0.8  & 460.5 ± 2.2 & 36.6 ± 0.6  & 507.8 ± 2.3 & 78.3 ± 2.8  & 480.4 ± 3.5 & 51.2 ± 1.2  \\   \toprule \bottomrule
\end{tabular}
}
\label{Runtime for dataset 1203_BNG_pwLinear}}
\end{table*}

\subsubsection{Memory (RAM) Usage Analysis}

As the cache size gradually increases, the demand for RAM size also rises accordingly. Therefore, we need to pay attention to RAM usage. Here, we provide a detailed comparison of the average RAM usage and maximum RAM usage under different cache size and cache strategies (see Tables \ref{RAM usage for dataset 344_mv}, \ref{RAM usage for dataset 215_2dplanes}, \ref{RAM usage for dataset 1203_BNG_pwLinear}).
We use the memory\_profiler library to test the maximum and average memory usage of gplearn, and all results are measured in megabytes (MB).

We can observe that both the average RAM usage and the peak RAM usage almost increase proportionally with the cache size, regardless of the cache strategy chosen. This indicates that the RAM consumption is highly sensitive to the cache size, suggesting a linear relationship between them as expected.

\begin{figure}[!t]
  \centering
  \includegraphics[width=\linewidth]{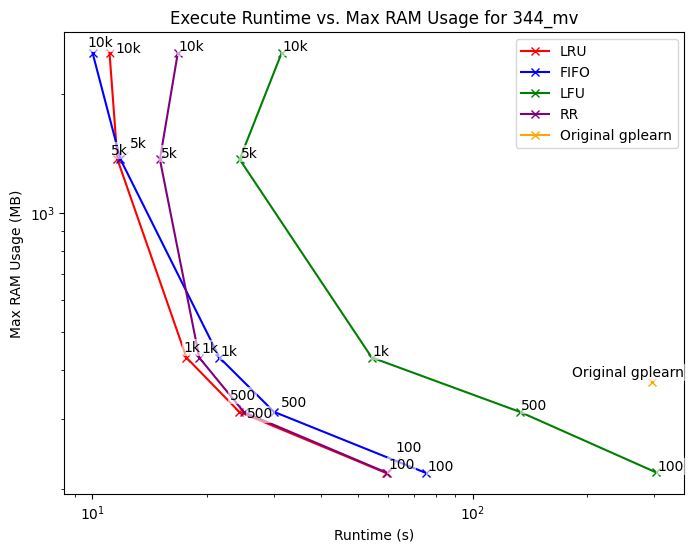}
  \vspace{-8mm}
  \caption{Impact of cache size (100, 500, 1k, 5k, 10k) on \textit{execute} runtime and maximum RAM usage across different cache strategies for dataset 344\_mv.}
  \label{runtime_RAM_344_mv}
  \vspace{-4mm}
\end{figure}

It is worth noting that the RAM usage of 50k and 100k is almost the same in dataset 344\_mv. This is because the fitting result is too simple and all the expressions cannot fill the 100k cache. Therefore, the cache sizes of 50k and 100k cannot show the effect of different cache strategies, and we will not show these two sets of data in Figure \ref{runtime_RAM_344_mv}.
Also in Tables \ref{Runtime for dataset 1203_BNG_pwLinear} and \ref{RAM usage for dataset 1203_BNG_pwLinear}, we did not test the results of the 1203\_BNG\_pwLinear dataset with cache sizes of 50k and 100k. This is primarily for the fitting results of the 1203\_BNG\_pwLinear dataset are similar in length to those of the 344\_mv dataset. The results are not complex, and Table \ref{Runtime for dataset 344_mv} shows that further increasing the cache size does not lead to significant improvements. Moreover, since the 1203\_BNG\_pwLinear dataset has a significantly larger number of samples compared to the other two datasets, the RAM requirements for the same cache size are much higher, exceeding the maximum RAM capacity of our server. For these reasons, we did not test this part of the data.

\begin{figure}[!t]
  \centering
  \includegraphics[width=\linewidth]{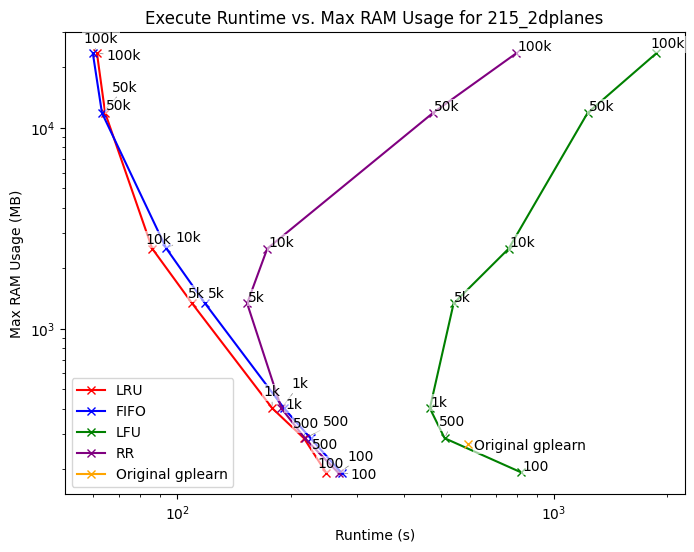}
  \vspace{-8mm}
  \caption{Impact of cache size (100, 500, 1k, 5k, 10k, 50k, 100k) on \textit{execute} runtime and maximum RAM usage across different cache strategies for dataset 215\_2dplanes.}
  \label{runtime_RAM_215_2dplanes}
  \vspace{-4mm}
\end{figure}

\begin{figure}[!t]
  \centering
  \includegraphics[width=\linewidth]{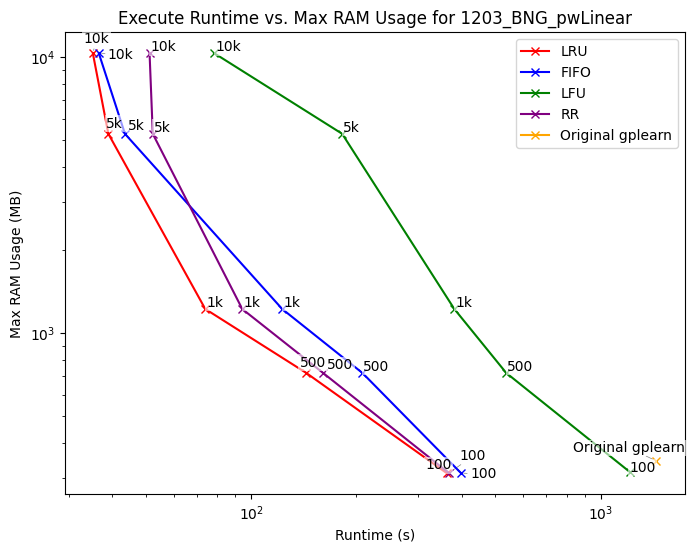}
  \vspace{-8mm}
  \caption{Impact of cache size (100, 500, 1k, 5k, 10k) on \textit{execute} runtime and maximum RAM usage across different cache strategies for dataset 1203\_BNG\_pwLinear.}
  \label{runtime_RAM_1203_BNG_pwLinear}
  \vspace{-4mm}
\end{figure}

Figures \ref{runtime_RAM_344_mv}, \ref{runtime_RAM_215_2dplanes} and \ref{runtime_RAM_1203_BNG_pwLinear}
provide a detailed comparison of different caching strategies across multiple datasets under various cache sizes, focusing on the cumulative runtime of the \textit{execute} function and the maximum RAM usage. Points closer to the lower-left corner in the plots indicate better overall performance. To better highlight the differences between caching strategies, logarithmic axes were used in the plots.

It can be observed that the LFU strategy is not well-suited for caching in gplearn. LRU generally outperforms FIFO in most cases. Although FIFO slightly reduces more runtime than LRU in some scenarios, this typically occurs when the cache size is large enough to significantly lower the \textit{execute} runtime, making the difference relatively insignificant in terms of the overall runtime. Considering the substantial RAM usage required by a 100k cache size and its limited impact on runtime optimization, even when the fitting result length reaches 100, such a large cache size appears relatively unnecessary.

It is also important to note that in these figures, the original gplearn calculates out of bag (OOB) fitness regardless of whether all max\_sample is set to 1, which unnecessarily consumes some RAM. In our modified version of gplearn, we always set max\_sample to the default value of 1, so we removed this portion of the code. As a result, the original gplearn uses even more memory than the gplearn with a cache size of 100.

\begin{table*}[!t]
\fontsize{9pt}{9.5pt}\selectfont
\centering
\caption{RAM Usage (Measured in MB) for Dataset 344\_mv}
\resizebox{\textwidth}{!}{
\begin{tabular}{l|ll|ll|ll|ll}
\bottomrule
~ & \multicolumn{2}{c|}{LRU} & \multicolumn{2}{c|}{FIFO} & \multicolumn{2}{c|}{LFU} & \multicolumn{2}{c}{RR} \\
Cache size & Average & Maximum & Average & Maximum & Average & Maximum & Average & Maximum  \\ \toprule \bottomrule
100 & 178.9 ± 8.4 & 219.4 ± 13.0  & 178.8 ± 8.3 & 219.5 ± 12.9  & 178.8 ± 7.7 & 220.2 ± 12.9  & 179.0 ± 8.3 & 219.5 ± 12.9  \\ 
500 & 273.0 ± 8.2 & 313.6 ± 12.8  & 273.0 ± 8.2 & 313.5 ± 12.9  & 272.3 ± 8.0 & 313.2 ± 12.9  & 272.6 ± 8.2 & 312.9 ± 12.9  \\ 
1k & 389.8 ± 8.2 & 430.0 ± 13.0  & 389.7 ± 8.3 & 430.0 ± 12.9  & 389.1 ± 6.9 & 430.3 ± 13.1  & 389.5 ± 8.1 & 429.7 ± 12.9  \\ 
5k & 1321.6 ± 8.1 & 1365.0 ± 12.9  & 1321.7 ± 8.1 & 1365.1 ± 13.1  & 1321.3 ± 7.6 & 1364.8 ± 13.0  & 1320.9 ± 8.0 & 1364.6 ± 12.8  \\ 
10k & 2482.1 ± 9.1 & 2534.4 ± 12.8  & 2479.3 ± 7.9 & 2534.4 ± 12.6  & 2480.1 ± 7.3 & 2534.0 ± 13.0  & 2478.5 ± 7.9 & 2533.2 ± 12.8  \\ 
50k & 8857.2 ± 369.3 & 11736.3 ± 168.2  & 8858.4 ± 369.3 & 11736.4 ± 168.2  & 9664.2 ± 138.5 & 11733.9 ± 168.2  & 9183.7 ± 234.3 & 11731.4 ± 168.1  \\ 
100k & 9151.2 ± 184.2 & 12454.0 ± 901.8  & 9150.4 ± 184.7 & 12454.0 ± 901.6  & 9962.6 ± 250.8 & 12451.5 ± 901.6  & 9413.7 ± 47.9 & 12449.1 ± 901.6  \\  \toprule \bottomrule
\end{tabular}
}
\label{RAM usage for dataset 344_mv}
\end{table*}

\begin{table*}[!t]
\fontsize{9pt}{9.5pt}\selectfont
\centering
\caption{RAM Usage (Measured in MB) for Dataset 215\_2dplanes}
\resizebox{\textwidth}{!}{
\begin{tabular}{l|ll|ll|ll|ll}
\bottomrule
~ & \multicolumn{2}{c|}{LRU} & \multicolumn{2}{c|}{FIFO} & \multicolumn{2}{c|}{LFU} & \multicolumn{2}{c}{RR} \\
Cache size & Average & Maximum & Average & Maximum & Average & Maximum & Average & Maximum \\ \toprule \bottomrule
100 & 171.2 ± 0.8 & 191.2 ± 1.5  & 171.3 ± 1.0 & 191.5 ± 1.3  & 171.8 ± 0.6 & 192.7 ± 0.7  & 171.4 ± 1.0 & 191.2 ± 1.3  \\ 
500 & 265.0 ± 0.8 & 285.1 ± 1.7  & 265.0 ± 0.9 & 285.2 ± 1.5  & 264.6 ± 0.9 & 285.3 ± 0.8  & 264.6 ± 1.1 & 284.6 ± 1.3  \\ 
1k & 381.6 ± 1.3 & 402.1 ± 1.6  & 381.7 ± 1.1 & 402.1 ± 1.5  & 381.7 ± 0.6 & 402.0 ± 1.1  & 381.3 ± 1.3 & 401.7 ± 1.1  \\ 
5k & 1316.4 ± 1.4 & 1338.8 ± 0.5  & 1319.4 ± 1.5 & 1343.7 ± 1.8  & 1315.7 ± 1.9 & 1336.8 ± 1.3  & 1315.6 ± 1.7 & 1337.9 ± 0.5  \\ 
10k & 2477.6 ± 3.1 & 2509.7 ± 1.5  & 2479.5 ± 4.6 & 2514.2 ± 0.9  & 2482.8 ± 1.9 & 2506.1 ± 1.4  & 2478.9 ± 4.0 & 2508.2 ± 1.5  \\ 
50k & 11347.5 ± 69.4 & 11882.4 ± 6.3  & 11356.0 ± 61.6 & 11881.6 ± 5.7  & 11741.7 ± 76.0 & 11875.9 ± 3.2  & 11641.3 ± 98.5 & 11871.0 ± 5.6  \\ 
100k & 20459.2 ± 560.7 & 23581.3 ± 23.5  & 20473.2 ± 549.0 & 23582.3 ± 23.3  & 22984.7 ± 454.3 & 23571.3 ± 17.3  & 22451.4 ± 815.5 & 23563.7 ± 16.8  \\ \toprule \bottomrule
\end{tabular}
}
\label{RAM usage for dataset 215_2dplanes}
\end{table*}

\begin{table*}[!t]
\fontsize{9pt}{9.5pt}\selectfont
\centering
\caption{RAM Usage (Measured in MB) for Dataset 1203\_BNG\_pwLinear}
\vspace{-4mm}
\resizebox{\textwidth}{!}{
\begin{tabular}{l|ll|ll|ll|ll}
\bottomrule
~ & \multicolumn{2}{c|}{LRU} & \multicolumn{2}{c|}{FIFO} & \multicolumn{2}{c|}{LFU}  & \multicolumn{2}{c}{RR} \\
Cache size & Average & Maximum & Average & Maximum & Average & Maximum & Average & Maximum  \\ \toprule \bottomrule
100 & 287.4 ± 0.3 & 311.6 ± 0.2  & 287.6 ± 0.2 & 311.7 ± 0.3  & 291.1 ± 0.1 & 314.9 ± 0.3  & 288.0 ± 0.1 & 311.8 ± 0.4  \\ 
500 & 693.2 ± 0.3 & 718.3 ± 0.3  & 693.4 ± 0.3 & 718.5 ± 0.3  & 694.8 ± 0.3 & 719.0 ± 0.3  & 692.8 ± 0.4 & 717.6 ± 0.5  \\ 
1k & 1198.9 ± 0.2 & 1225.3 ± 0.3  & 1199.2 ± 0.3 & 1225.3 ± 0.4  & 1200.7 ± 0.1 & 1225.8 ± 0.4  & 1198.6 ± 0.1 & 1224.6 ± 0.1  \\ 
5k & 5230.6 ± 0.4 & 5281.5 ± 0.3  & 5230.8 ± 0.4 & 5281.3 ± 0.2  & 5237.7 ± 0.1 & 5282.2 ± 0.2  & 5230.5 ± 0.3 & 5281.1 ± 0.2  \\ 
10k & 10225.5 ± 0.6 & 10352.4 ± 0.3  & 10225.4 ± 0.6 & 10352.3 ± 0.3  & 10234.4 ± 0.7 & 10351.5 ± 0.6  & 10227.2 ± 0.4 & 10351.1 ± 0.1  \\   \toprule \bottomrule
\end{tabular}
}
\label{RAM usage for dataset 1203_BNG_pwLinear}
\vspace{-3mm}
\end{table*}

\subsection{Proactive Periodic Cache Clearance}

We also investigate whether clearing the cache completely once every $n$ generations can help reduce the training time of gplearn. In this section, we use a large synthetic dataset with 10,000 samples and set the population size to 10,000. The ground truth target function is defined as $y_{\text{truth}} = x_0^3 + x_1^2 + x_0 + \sin(x_1) + \sin(x_0^2)$.

Gplearn is trained for 15 generations and we actively clear the cache every 1, 2, 3, 4, and 5 generations. We also have a control group where the cache is never cleared to analyze whether there is a significant difference between the overall training runtime and the \textit{execute} function runtime under these different conditions.

As shown in Table \ref{Regularly clear cache}, we find that regardless of how we adjust the frequency of actively clearing the cache, it has almost no statistically significant effect on the training time of gplearn, whether in terms of the overall runtime or the cumulative runtime of the \textit{execute} function. When the cache becomes full and new key-value pairs are recorded, the cache will delete the oldest key-value pairs based on predefined rules. These results suggest that there may be no need to actively clean the cache since the time saved from not having to do frequent deletions is not worth it.

\begin{table}[!t]
\fontsize{9pt}{9.5pt}\selectfont
\centering
\caption{Proactive Periodic Cache Clearance}
\vspace{-3mm}
\begin{tabular}{l|ll}
\bottomrule
Cache Clearing& Overall & \textit{execute}\\
Frequency (Generations)& \\
\toprule \bottomrule
1      & 79.21 & 17.75 \\
2      & 79.71 & 18.05 \\
3      & 80.10 & 18.23 \\
4      & 79.21 & 18.10 \\
5      & 79.73 & 18.16 \\
Never  & 80.54 & 18.48 \\ \toprule \bottomrule
\end{tabular}
\label{Regularly clear cache}
\vspace{-2mm}
\end{table}

\subsection{Access Count Test for Infinite Cache}

In this section, we conducted experiments using hyperparameters that are nearly identical to those used in the preceding subsection. The only difference is that we implemented an infinitely large cache using a Python dictionary, which allowed us to record the call count for each cached key-value pair. This feature facilitated the analysis and selection of an appropriate cache size.

For the infinite cache size scenario, the cache was invoked a total of 273,553 times. Analysis revealed that the top 6,070 most frequently accessed items accounted for 90\% of the total calls, while representing only 2.41\% of the entire cache size. Furthermore, the top 5,000 items alone accounted for 203,305 calls, which is 87.87\% of the total. Notably, the runtime improvements nearly diminished beyond a cache size of 5,000, underscoring the practicality of this threshold.

\subsection{Another Memory-Runtime Trade-off}\label{sect:tradeoff}

In the field of computer science, core hour is often used as a metric for measuring computing resource usage, particularly in high-performance computing (HPC) environments. The formula for calculating core hours is:
\[
\text{Core Hours} = \text{Number of Cores} \times \text{Execution Time (hours)}
\]
For example, using 10 cores for 1 hour consumes 10 core hours. Similarly, using 1 core for 10 hours also consumes 10 core hours.

Inspired by this evaluation metric, we propose to use the following metric to analyze GPSR: RAM hour, defined as:
\[
\text{RAM Hour} = \text{Max RAM Usage (GB)} \times \text{Execution Time (hours)}
\]

The RAM hour metric offers a comprehensive evaluation of the caching strategy by simultaneously considering two crucial factors: the reduction in execution time caused by caching and the additional RAM usage introduced by caching. The RAM hour is not an arbitrary formulation, but instead directly captures the trade-off between memory and time, in which the settings with the least RAM hour can be shown to be most efficient use of resources.  

\begin{figure}[!t]
  \centering
  \includegraphics[width=0.9\linewidth]{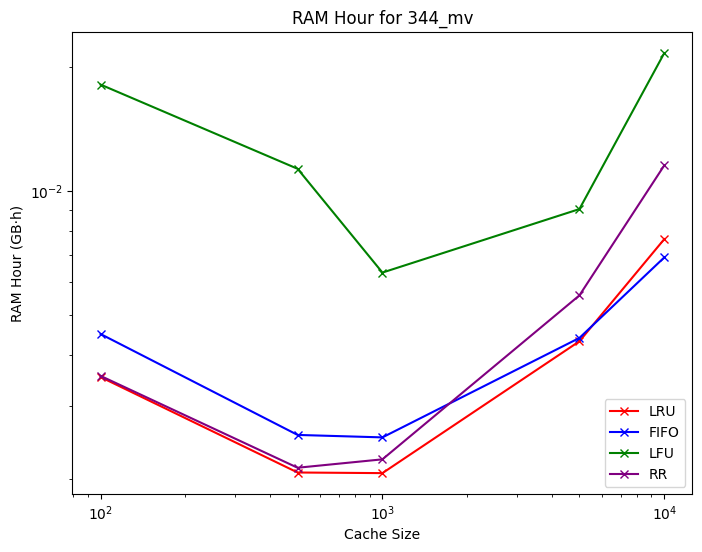}
  \vspace{-4mm}
  \caption{RAM Hour for Dataset 344\_mv}
  \label{ramhour_344_mv}
  \vspace{-4mm}
\end{figure}

\begin{figure}[!t]
  \centering
  \includegraphics[width=0.9\linewidth]{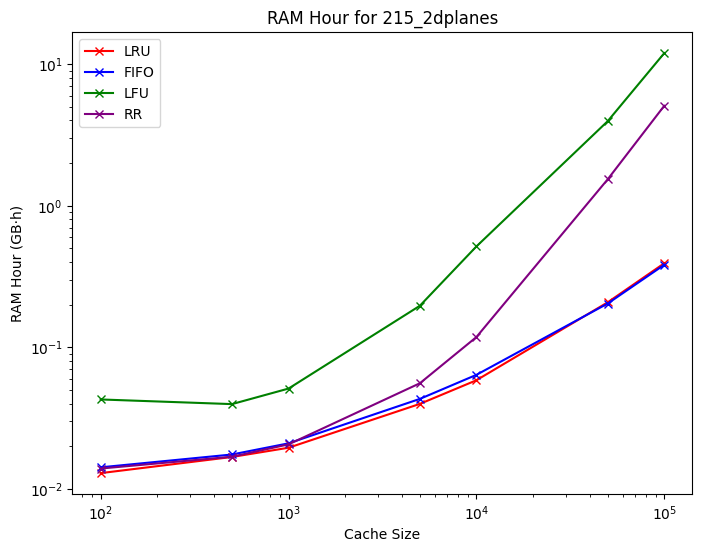}
  \vspace{-4mm}
  \caption{RAM Hour for Dataset 215\_2dplanes}
  \label{ramhour_215_2dplanes}
  \vspace{-4mm}
\end{figure}

\begin{figure}[!t]
  \centering
  \includegraphics[width=0.9\linewidth]{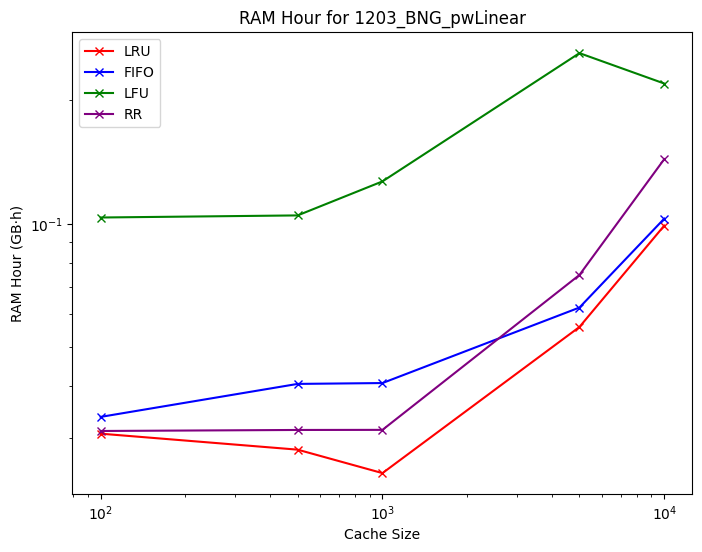}
  \vspace{-4mm}
  \caption{RAM Hour for Dataset 1203\_BNG\_pwLinear}
  \label{ramhour_1203_BNG_pwLinear}
  \vspace{-4mm}
\end{figure}

High RAM usage typically increases the cost of renting a server per hour, while long execution times can lead to higher server rental times. The RAM-hour metric encapsulates these factors into a single scalar value, making it an accurate and meaningful way to quantify efficiency in caching strategies. 

Thus, its adoption is not merely a heuristic choice but an approach that aligns closely with the resource allocation and optimization. By minimizing RAM hour, one can optimize resource utilization under realistic system constraints.

Figures \ref{ramhour_344_mv}, \ref{ramhour_215_2dplanes}, and \ref{ramhour_1203_BNG_pwLinear} illustrate the RAM Hour of three different datasets—344\_mv, 215\_2dplanes, and 1203\_BNG\_pwLinear—under different cache sizes. A lower RAM Hour indicates better overall performance of the cache.

From the results, we observe that for the datasets 344\_mv and 1203\_BNG\_pwLinear, the RAM Hour is lowest when using the LRU strategy with a cache size of 1,000, demonstrating the best overall performance for these two datasets under this cache size configuration. For the 215\_2dplanes dataset, the RAM Hour generally increases with larger cache sizes. However, this does not imply that setting the cache size to 100 is optimal, as an excessively small cache size may fail to provide sufficient acceleration. Notably, the rate of increase in RAM Hour is not uniform: before the cache size reaches 1,000, the RAM Hour rises relatively slowly, after which the increase becomes more pronounced. Thus, from the perspective of the slope of the RAM Hour curve, a cache size of 1,000 appears to be a balanced choice for the 215\_2dplanes dataset.

\subsection{Guidelines for Appropriate Cache Size}

\underline{Case 1:} If efficient use of resources is not of concern, and even RAM size is unlimited, then from the analysis in the previous section, we know that the runtime of GPSR still decreases, albeit gradually, as the cache size increases for most cache strategies. After a certain cache size threshold, the improvement in runtime becomes increasingly limited. In some caching strategies, such as LFU, a larger cache size may even result in longer runtimes. Fortunately, with an effective caching strategy, such as LRU or FIFO, excessively large caches do not have a significant adverse effect on runtime. Therefore, if there are no strict constraints on RAM usage, relatively larger cache sizes, such as 10,000 to 100,000 on LRU or FIFO should be used since the runtime will still indeed decrease.

\noindent\underline{Case 2:} If the efficient use of resources is important or if there is limited memory, the appropriate cache size is dependent on the use-case of GPSR. We provide guidelines for 2 separate cases: 1). Sequential GPSR experiments (e.g., algorithm development, preliminary research), 2). Parallel GPSR experiments (e.g., gathering statistically significant results by running the same experiments over multiple similar-complexity datasets or random seeds).

\noindent\underline{Case 3:} In the case where GPSR experiments are run sequentially (such as when doing algorithm development and preliminary research), at any time, only one instance of GPSR is using the RAM. In that case, practitioners should analyze the aspect of dataset complexity. To illustrate, consider the 344\_mv dataset. When the dataset is partitioned with a random state of 1 and using LRU cache strategy, the fitting result is y = div(add(X5, X0), div(X8, X7)).
The expression length is relatively short, selecting a cache size of 5,000 already can reduce the runtime of the \textit{execute} function significantly, decreasing it from 302.4 seconds to 11.1 seconds. Further increasing the cache size does not substantially improve the runtime. However, under the same conditions, when we switch the dataset to 215\_2dplanes, the fitting result becomes extremely complex and long:
y = add(add(X0, add(X3, X5)), add(add(X2, X1), add(add(div(add(X0, add(X3, X0)), add(add(X0, X0), add(X4, X0))), add(div(add(mul(-0.388, add(add(X5, add(X5, add(X0, add(add(X5, add(X4, X0)), add(X4, add(div(add(X3, mul(-0.388, add(add(add(X4, X6), add(add(add(X1, X6), add(add(X4, X6), add(div(add(X3, X5), X0), X1))), X1)), X4))), X0), X1)))))), add(X4, X6))), X1), X0), X0)), add(mul(add(X2, X1), X0), add(X4, X0))))).
Therefore, even when increasing the cache size to 10,000, the runtime of the \textit{execute} function is only reduced from 1261.3 seconds to 415.5 seconds. Selecting a larger cache size still provides some improvement in runtime. If we continue to increase the cache size to 50,000, the \textit{execute} runtime will continue to decrease to 214.5 seconds. Since the complexity of real-world dataset is often not something we can explicitly quantify a priori, we recommend that the best practice would be to perform a quick preliminary analysis to generate memory-runtime plots like Figures \ref{runtime_RAM_344_mv}, \ref{runtime_RAM_215_2dplanes}, \ref{runtime_RAM_1203_BNG_pwLinear} to determine this point using suitable methods like the elbow method.

\noindent\underline{Case 4:} In the case where GPSR experiments are run in parallel, such as when gathering statistically significant results by running the same experiments over multiple similar-complexity datasets or random seeds, then the approach is different. We recommend to conduct the same preliminary experiment as above, but also plot RAM hour against cache size (see Figures \ref{ramhour_344_mv}, \ref{ramhour_215_2dplanes}, \ref{ramhour_1203_BNG_pwLinear}). The optimal point will be the settings with the lowest RAM hour. Then, the recommended number of GPSR to be run in parallel is the maximum memory of the server, divided by the max RAM usage at the settings of the optimal point. By running this number of GPSR in parallel, it will encourage the server to operate at the optimal point. At this point, the server's memory will be at 100\% utilization rate and require the shortest total runtime compared to any other strategy.

\section{CONCLUSION}

In this work, we extend GPSR by incorporating a variety of caching mechanism to accelerate the computation time of fitness evaluation. Conducting a variety of experiments on PMLB datasets, we comprehensively compared the performance impacts and memory-runtime trade-offs of different caching strategies and cache sizes. Our experimental results demonstrate that selecting an appropriate caching strategy and cache size can significantly reduce the computation time for fitness evaluation, achieving improvements by an order of magnitude or even more. Additionally, we investigate whether periodic proactive cache cleaning could further enhance cache performance and analyzed the frequency of item accesses in the cache under the assumption of an unlimited cache size. Then, we propose utilizing the RAM hour metric to capture the trade-off between memory and time in a straightforward and meaningful way. These analyses provide strong validation and culminated in our proposed recommendations for optimizing cache strategies and configurations for GPSR users and researchers.

\section*{ACKNOWLEDGMENTS}
This research/project is supported by the National Research Foundation, Singapore under its AI Singapore Programme (AISG Award No: AISG3-PhD-2023-08-052T), and A*STAR, CISCO Systems (USA) Pte. Ltd and National University of Singapore under its Cisco-NUS Accelerated Digital Economy Corporate Laboratory (Award I21001E0002).


\bibliographystyle{ACM-Reference-Format}
\bibliography{references}


\begin{thebibliography}{25}


\ifx \showCODEN    \undefined \def \showCODEN     #1{\unskip}     \fi
\ifx \showISBNx    \undefined \def \showISBNx     #1{\unskip}     \fi
\ifx \showISBNxiii \undefined \def \showISBNxiii  #1{\unskip}     \fi
\ifx \showISSN     \undefined \def \showISSN      #1{\unskip}     \fi
\ifx \showLCCN     \undefined \def \showLCCN      #1{\unskip}     \fi
\ifx \shownote     \undefined \def \shownote      #1{#1}          \fi
\ifx \showarticletitle \undefined \def \showarticletitle #1{#1}   \fi
\ifx \showURL      \undefined \def \showURL       {\relax}        \fi
\providecommand\bibfield[2]{#2}
\providecommand\bibinfo[2]{#2}
\providecommand\natexlab[1]{#1}
\providecommand\showeprint[2][]{arXiv:#2}

\bibitem[Ahmadi~Daryakenari et~al\mbox{.}(2024)]%
        {ahmadi2024ai}
\bibfield{author}{\bibinfo{person}{Nazanin Ahmadi~Daryakenari}, \bibinfo{person}{Mario De~Florio}, \bibinfo{person}{Khemraj Shukla}, {and} \bibinfo{person}{George~Em Karniadakis}.} \bibinfo{year}{2024}\natexlab{}.
\newblock \showarticletitle{AI-Aristotle: A physics-informed framework for systems biology gray-box identification}.
\newblock \bibinfo{journal}{\emph{PLOS Computational Biology}} \bibinfo{volume}{20}, \bibinfo{number}{3} (\bibinfo{year}{2024}), \bibinfo{pages}{e1011916}.
\newblock


\bibitem[Bakurov et~al\mbox{.}(2022)]%
        {bakurov2022genetic}
\bibfield{author}{\bibinfo{person}{Illya Bakurov}, \bibinfo{person}{Marco Buzzelli}, \bibinfo{person}{Mauro Castelli}, \bibinfo{person}{Raimondo Schettini}, {and} \bibinfo{person}{Leonardo Vanneschi}.} \bibinfo{year}{2022}\natexlab{}.
\newblock \showarticletitle{Genetic programming for structural similarity design at multiple spatial scales}. In \bibinfo{booktitle}{\emph{Proceedings of the Genetic and Evolutionary Computation Conference}}. \bibinfo{pages}{911--919}.
\newblock


\bibitem[Banzhaf and Bakurov(2024)]%
        {banzhaf2024nature}
\bibfield{author}{\bibinfo{person}{Wolfgang Banzhaf} {and} \bibinfo{person}{Illya Bakurov}.} \bibinfo{year}{2024}\natexlab{}.
\newblock \showarticletitle{On the Nature of the Phenotype in Tree Genetic Programming}. In \bibinfo{booktitle}{\emph{Proceedings of the Genetic and Evolutionary Computation Conference}}. \bibinfo{pages}{868--877}.
\newblock


\bibitem[Belady(1966)]%
        {belady1966study}
\bibfield{author}{\bibinfo{person}{Laszlo~A. Belady}.} \bibinfo{year}{1966}\natexlab{}.
\newblock \showarticletitle{A study of replacement algorithms for a virtual-storage computer}.
\newblock \bibinfo{journal}{\emph{IBM Systems journal}} \bibinfo{volume}{5}, \bibinfo{number}{2} (\bibinfo{year}{1966}), \bibinfo{pages}{78--101}.
\newblock


\bibitem[Chen et~al\mbox{.}(2021)]%
        {chen2021empirical}
\bibfield{author}{\bibinfo{person}{Tianxiang Chen}, \bibinfo{person}{Wei Chen}, {and} \bibinfo{person}{Luyao Du}.} \bibinfo{year}{2021}\natexlab{}.
\newblock \showarticletitle{An Empirical Study of Financial Factor Mining Based on Gene Expression Programming}. In \bibinfo{booktitle}{\emph{2021 4th International Conference on Advanced Electronic Materials, Computers and Software Engineering (AEMCSE)}}. IEEE, \bibinfo{pages}{1113--1117}.
\newblock


\bibitem[Cranmer(2023)]%
        {cranmer2023interpretable}
\bibfield{author}{\bibinfo{person}{Miles Cranmer}.} \bibinfo{year}{2023}\natexlab{}.
\newblock \showarticletitle{Interpretable machine learning for science with PySR and SymbolicRegression. jl}.
\newblock \bibinfo{journal}{\emph{arXiv preprint arXiv:2305.01582}} (\bibinfo{year}{2023}).
\newblock


\bibitem[de~Franca and Kronberger(2023)]%
        {de2023reducing}
\bibfield{author}{\bibinfo{person}{Fabricio~Olivetti de Franca} {and} \bibinfo{person}{Gabriel Kronberger}.} \bibinfo{year}{2023}\natexlab{}.
\newblock \showarticletitle{Reducing Overparameterization of Symbolic Regression Models with Equality Saturation}. In \bibinfo{booktitle}{\emph{Proceedings of the Genetic and Evolutionary Computation Conference}}. \bibinfo{pages}{1064--1072}.
\newblock


\bibitem[Downey and Zhang(2011)]%
        {5949751}
\bibfield{author}{\bibinfo{person}{Carlton Downey} {and} \bibinfo{person}{Mengjie Zhang}.} \bibinfo{year}{2011}\natexlab{}.
\newblock \showarticletitle{Execution trace caching for Linear Genetic Programming}. In \bibinfo{booktitle}{\emph{2011 IEEE Congress of Evolutionary Computation (CEC)}}. \bibinfo{pages}{1186--1193}.
\newblock
\href{https://doi.org/10.1109/CEC.2011.5949751}{doi:\nolinkurl{10.1109/CEC.2011.5949751}}


\bibitem[Fong and Motani(2024a)]%
        {fong2024metasr}
\bibfield{author}{\bibinfo{person}{Kei~Sen Fong} {and} \bibinfo{person}{Mehul Motani}.} \bibinfo{year}{2024}\natexlab{a}.
\newblock \showarticletitle{MetaSR: A Meta-Learning Approach to Fitness Formulation for Frequency-Aware Symbolic Regression}. In \bibinfo{booktitle}{\emph{Proceedings of the Genetic and Evolutionary Computation Conference}}. \bibinfo{pages}{878--886}.
\newblock


\bibitem[Fong and Motani(2024b)]%
        {fong2024multi}
\bibfield{author}{\bibinfo{person}{Kei~Sen Fong} {and} \bibinfo{person}{Mehul Motani}.} \bibinfo{year}{2024}\natexlab{b}.
\newblock \showarticletitle{Multi-level symbolic regression: Function structure learning for multi-level data}. In \bibinfo{booktitle}{\emph{International conference on artificial intelligence and statistics}}. PMLR, \bibinfo{pages}{2890--2898}.
\newblock


\bibitem[Fong and Motani(2024c)]%
        {fong2024symbolic}
\bibfield{author}{\bibinfo{person}{Kei~Sen Fong} {and} \bibinfo{person}{Mehul Motani}.} \bibinfo{year}{2024}\natexlab{c}.
\newblock \showarticletitle{Symbolic regression enhanced decision trees for classification tasks}. In \bibinfo{booktitle}{\emph{Proceedings of the AAAI Conference on Artificial Intelligence}}, Vol.~\bibinfo{volume}{38}. \bibinfo{pages}{12033--12042}.
\newblock


\bibitem[Geiger et~al\mbox{.}(2023)]%
        {geiger2023down}
\bibfield{author}{\bibinfo{person}{Alina Geiger}, \bibinfo{person}{Dominik Sobania}, {and} \bibinfo{person}{Franz Rothlauf}.} \bibinfo{year}{2023}\natexlab{}.
\newblock \showarticletitle{Down-sampled epsilon-lexicase selection for real-world symbolic regression problems}. In \bibinfo{booktitle}{\emph{Proceedings of the Genetic and Evolutionary Computation Conference}}. \bibinfo{pages}{1109--1117}.
\newblock


\bibitem[Handley(1994)]%
        {350024}
\bibfield{author}{\bibinfo{person}{S. Handley}.} \bibinfo{year}{1994}\natexlab{}.
\newblock \showarticletitle{On the use of a directed acyclic graph to represent a population of computer programs}. In \bibinfo{booktitle}{\emph{Proceedings of the First IEEE Conference on Evolutionary Computation. IEEE World Congress on Computational Intelligence}}. \bibinfo{pages}{154--159 vol.1}.
\newblock
\href{https://doi.org/10.1109/ICEC.1994.350024}{doi:\nolinkurl{10.1109/ICEC.1994.350024}}


\bibitem[He et~al\mbox{.}(2022)]%
        {he2022taylor}
\bibfield{author}{\bibinfo{person}{Baihe He}, \bibinfo{person}{Qiang Lu}, \bibinfo{person}{Qingyun Yang}, \bibinfo{person}{Jake Luo}, {and} \bibinfo{person}{Zhiguang Wang}.} \bibinfo{year}{2022}\natexlab{}.
\newblock \showarticletitle{Taylor genetic programming for symbolic regression}. In \bibinfo{booktitle}{\emph{Proceedings of the genetic and evolutionary computation conference}}. \bibinfo{pages}{946--954}.
\newblock


\bibitem[Imai~Aldeia et~al\mbox{.}(2024)]%
        {imai2024inexact}
\bibfield{author}{\bibinfo{person}{Guilherme~Seidyo Imai~Aldeia}, \bibinfo{person}{Fabr{\'\i}cio~Olivetti De~Fran{\c{c}}a}, {and} \bibinfo{person}{William~G La~Cava}.} \bibinfo{year}{2024}\natexlab{}.
\newblock \showarticletitle{Inexact Simplification of Symbolic Regression Expressions with Locality-sensitive Hashing}. In \bibinfo{booktitle}{\emph{Proceedings of the Genetic and Evolutionary Computation Conference}}. \bibinfo{pages}{896--904}.
\newblock


\bibitem[Jameer and Ajith(2024)]%
        {jameer2024stability}
\bibfield{author}{\bibinfo{person}{Basha~SK Jameer} {and} \bibinfo{person}{Jubilson~E Ajith}.} \bibinfo{year}{2024}\natexlab{}.
\newblock \showarticletitle{Stability Assessment of Mbene Materials Using Scikit-Learn Inspired Application Programming Interface: Gplearn and Particle Swarm Optimization}.
\newblock \bibinfo{journal}{\emph{ES Materials \& Manufacturing}}  \bibinfo{volume}{24} (\bibinfo{year}{2024}), \bibinfo{pages}{1166}.
\newblock


\bibitem[Keijzer(2004)]%
        {keijzer2004alternatives}
\bibfield{author}{\bibinfo{person}{Maarten Keijzer}.} \bibinfo{year}{2004}\natexlab{}.
\newblock \showarticletitle{Alternatives in subtree caching for genetic programming}. In \bibinfo{booktitle}{\emph{European Conference on Genetic Programming}}. Springer, \bibinfo{pages}{328--337}.
\newblock


\bibitem[Kemmer(2014)]%
        {cachetools}
\bibfield{author}{\bibinfo{person}{Thomas Kemmer}.} \bibinfo{year}{2014}\natexlab{}.
\newblock \bibinfo{title}{Cachetools: Extensible memoizing collections and decorators}.
\newblock \bibinfo{howpublished}{\url{https://github.com/tkem/cachetools}}.
\newblock


\bibitem[Koza(1992)]%
        {koza1992genetic}
\bibfield{author}{\bibinfo{person}{John~R Koza}.} \bibinfo{year}{1992}\natexlab{}.
\newblock \showarticletitle{Genetic programming. {O}n the programming of computers by means of natural selection}.
\newblock \bibinfo{journal}{\emph{Complex adaptive systems}} (\bibinfo{year}{1992}).
\newblock


\bibitem[Orzechowski et~al\mbox{.}(2018)]%
        {srbench}
\bibfield{author}{\bibinfo{person}{Patryk Orzechowski}, \bibinfo{person}{William La~Cava}, {and} \bibinfo{person}{Jason~H Moore}.} \bibinfo{year}{2018}\natexlab{}.
\newblock \showarticletitle{Where are we now? {A} large benchmark study of recent symbolic regression methods}. In \bibinfo{booktitle}{\emph{Proceedings of the Genetic and Evolutionary Computation Conference}}. \bibinfo{pages}{1183--1190}.
\newblock


\bibitem[Romano et~al\mbox{.}(2022)]%
        {romano2021pmlb}
\bibfield{author}{\bibinfo{person}{Joseph~D Romano}, \bibinfo{person}{Trang~T Le}, \bibinfo{person}{William La~Cava}, \bibinfo{person}{John~T Gregg}, \bibinfo{person}{Daniel~J Goldberg}, \bibinfo{person}{Praneel Chakraborty}, \bibinfo{person}{Natasha~L Ray}, \bibinfo{person}{Daniel Himmelstein}, \bibinfo{person}{Weixuan Fu}, {and} \bibinfo{person}{Jason~H Moore}.} \bibinfo{year}{2022}\natexlab{}.
\newblock \showarticletitle{PMLB v1. 0: an open-source dataset collection for benchmarking machine learning methods}.
\newblock \bibinfo{journal}{\emph{Bioinformatics}} \bibinfo{volume}{38}, \bibinfo{number}{3} (\bibinfo{year}{2022}), \bibinfo{pages}{878--880}.
\newblock


\bibitem[Smith(1987)]%
        {smith1987design}
\bibfield{author}{\bibinfo{person}{Alan~Jay Smith}.} \bibinfo{year}{1987}\natexlab{}.
\newblock \bibinfo{booktitle}{\emph{Design of CPU cache memories}}.
\newblock \bibinfo{publisher}{Computer Science Division, University of California}.
\newblock


\bibitem[Stephens(2016)]%
        {stephens2016genetic}
\bibfield{author}{\bibinfo{person}{Trevor Stephens}.} \bibinfo{year}{2016}\natexlab{}.
\newblock \showarticletitle{Genetic Programming in Python, with a scikit-learn inspired API: gplearn}.
\newblock \bibinfo{journal}{\emph{Python Library}} (\bibinfo{year}{2016}).
\newblock


\bibitem[Wong and Zhang(2008)]%
        {4631158}
\bibfield{author}{\bibinfo{person}{Phillip Wong} {and} \bibinfo{person}{Mengjie Zhang}.} \bibinfo{year}{2008}\natexlab{}.
\newblock \showarticletitle{SCHEME: Caching subtrees in genetic programming}. In \bibinfo{booktitle}{\emph{2008 IEEE Congress on Evolutionary Computation (IEEE World Congress on Computational Intelligence)}}. \bibinfo{pages}{2678--2685}.
\newblock
\href{https://doi.org/10.1109/CEC.2008.4631158}{doi:\nolinkurl{10.1109/CEC.2008.4631158}}


\bibitem[Wong(2008)]%
        {wong2008removing}
\bibfield{author}{\bibinfo{person}{Phillip Lee-Ming Wong}.} \bibinfo{year}{2008}\natexlab{}.
\newblock \emph{\bibinfo{title}{Removing redundancy and reducing fitness evaluation costs in genetic programming}}.
\newblock \bibinfo{thesistype}{Ph.\,D. Dissertation}. \bibinfo{school}{Open Access Te Herenga Waka-Victoria University of Wellington}.
\newblock


\end{thebibliography}

\appendix

\end{document}